\documentclass[10pt,twocolumn,letterpaper]{article}

\usepackage{iccv}
\usepackage{times}
\usepackage{epsfig}
\usepackage{booktabs}
\usepackage{graphicx}
\usepackage{graphics}
\usepackage{amsmath}
\usepackage{amssymb}
\usepackage{enumitem}
\usepackage{enumerate}
\usepackage{bm}
\usepackage{subfigure}
\usepackage{multirow}
\usepackage{array}
\usepackage{algorithm}
\usepackage{algpseudocode}

\usepackage[breaklinks=true,bookmarks=false]{hyperref}

\iccvfinalcopy 

\def\iccvPaperID{2669} 

\ificcvfinal\fi

\begin{document}

\title{Generalizable Mixed-Precision Quantization via Attribution Rank Preservation}

\author{Ziwei Wang\textsuperscript{1,2,3},
	Han Xiao\textsuperscript{1,2,3},
	Jiwen Lu\textsuperscript{1,2,3}\thanks{Corresponding author},
	Jie Zhou\textsuperscript{1,2,3}\\	
	\textsuperscript{1} Department of Automation, Tsinghua University, China\\
	\textsuperscript{2} State Key Lab of Intelligent Technologies and Systems, China\\
	\textsuperscript{3} Beijing National Research Center for Information Science and Technology, China\\
	{\tt \small \{wang-zw18, h-xiao20\}@mails.tsinghua.edu.cn; \{lujiwen,jzhou\}@tsinghua.edu.cn}
}

\maketitle
\ificcvfinal\thispagestyle{empty}\fi

\begin{abstract}
In this paper, we propose a generalizable mixed-precision quantization (GMPQ) method for efficient inference. Conventional methods require the consistency of datasets for bitwidth search and model deployment to guarantee the policy optimality, leading to heavy search cost on challenging largescale datasets in realistic applications. On the contrary, our GMPQ searches the mixed-quantization policy that can be generalized to largescale datasets with only a small amount of data, so that the search cost is significantly reduced without performance degradation. Specifically, we observe that locating network attribution correctly is general ability for accurate visual analysis across different data distribution. Therefore, despite of pursuing higher model accuracy and complexity, we preserve attribution rank consistency between the quantized models and their full-precision counterparts via efficient capacity-aware attribution imitation for generalizable mixed-precision quantization strategy search. Extensive experiments show that our method obtains competitive accuracy-complexity trade-off compared with the state-of-the-art mixed-precision networks in significantly reduced search cost. The code is available at \href{https://github.com/ZiweiWangTHU/GMPQ.git}{https://github.com/ZiweiWangTHU/GMPQ.git.}
\end{abstract}

\begin{figure}[t]
	\centering
	\includegraphics[height=11.1cm, width=8cm]{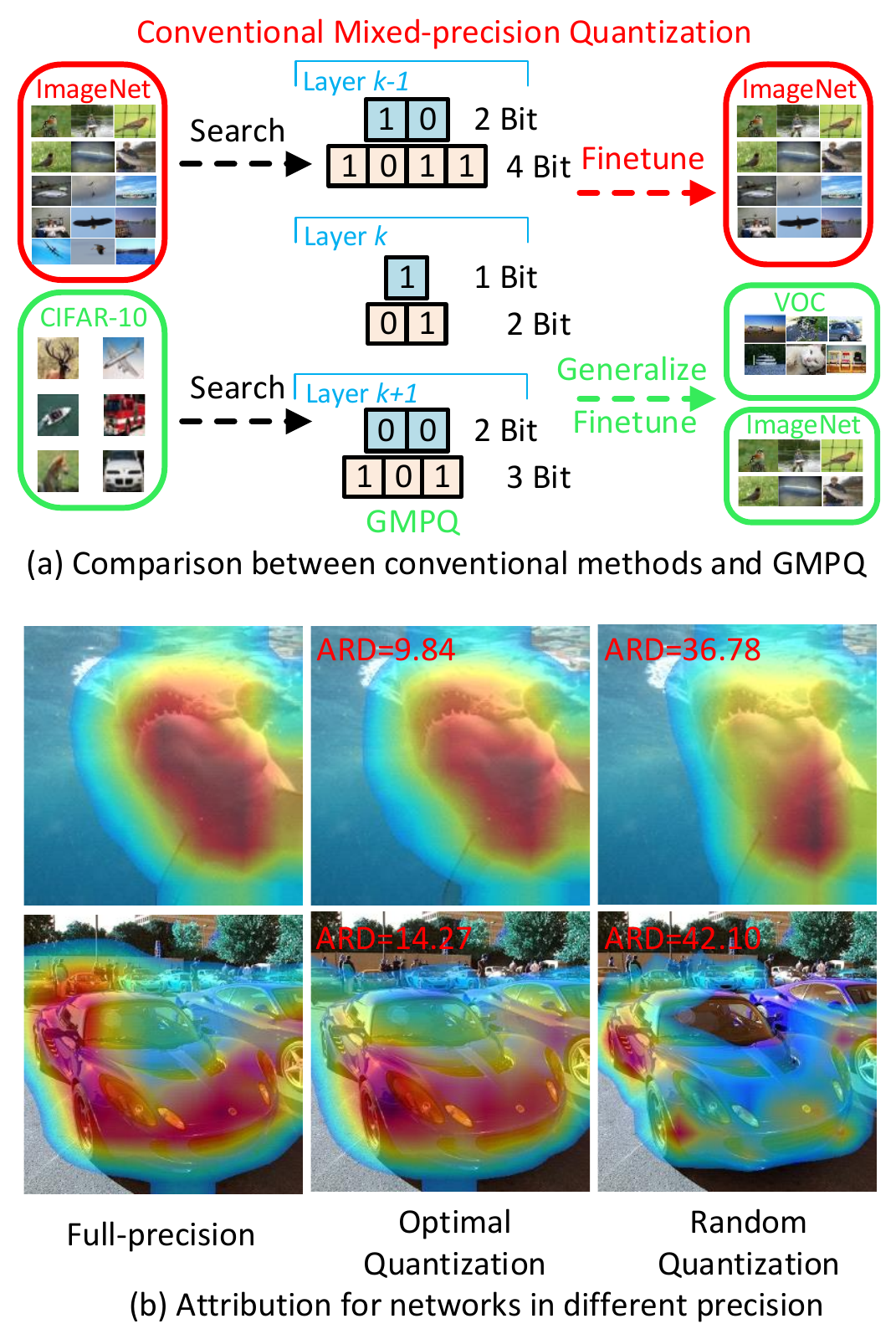}
	\caption{(a) Conventional methods require the consistency of datasets for bitwidth search and model deployment, while our GMPQ searches the optimal quantization policy on small datasets and generalizes it to largescale datasets. (b) The attribution computed by Grad-cam for images from ImageNet (top row) and PASCAL VOC (bottom row). Different from random quantization, the optimal quantization policy keeps similar attribution with the full-precision counterparts regardless of datasets. ARD means the average Attribution Rank Distance for the top-100 pixels with the highest attribution in the full-precision feature maps.}
	\vspace{-0.5cm}    
	\label{comparison}
\end{figure}

\section{Introduction}

Deep neural networks have achieved the state-of-the art performance across a large number of vision tasks such as image classification \cite{he2016deep,simonyan2014very,huang2017densely}, object detection \cite{ren2015faster,liu2016ssd,he2017mask}, face recognition \cite{deng2019arcface, wang2018cosface, liu2017sphereface} and many others. However, the mobile devices with limited storage and computational resources are not capable of processing deep models due to the extremely high complexity. Therefore, it is desirable to design network compression strategy according to the hardware configurations.

Recently, several network compression techniques have been proposed including pruning \cite{lin2017runtime, he2017channel, molchanov2019importance}, quantization \cite{zhao2019improving, liu2018bi, wang2020towards}, efficient architecture design \cite{iandola2016squeezenet, howard2017mobilenets, qin2019thundernet} and low-rank decomposition \cite{denton2014exploiting, yu2017compressing, li2020group}. Among these approaches, quantization constrains the network weights and activations in limited bitwidth for memory saving and fast processing. In order to fully utilize the hardware resources, mixed-precision quantization \cite{wang2019haq, dong2019hawq,cai2020rethinking} is presented to search the bitwidth in each layer so that the optimal accuracy-complexity trade-off is obtained. However, conventional mixed-precision quantization requires the consistency of datasets for bitwidth search and network deployment to guarantee policy optimality, which causes significant search burden for automated model compression on largescale datasets such as ImageNet \cite{deng2009imagenet}. For example, it usually takes several GPU days to acquire the expected quantization strategy for ResNet18 on ImageNet \cite{wang2019haq, cai2020rethinking}.

In this paper, we present a GMPQ method to learn generalizable mixed-precision quantization strategy via attribution rank preservation for efficient inference. Unlike existing methods which requires the dataset consistency between quantization policy search and model deployment, our method enables the acquired quantization strategy to be generalizable across various datasets. The quantization policy searched on small datasets achieves promising performance on challenging largescale datasets, so that policy search cost is significantly reduced. Figure \ref{comparison}(a) shows the difference between our GMPQ and conventional mixed-precision networks. More specifically, we observe that correctly locating the network attribution benefits visual analysis for various input data distribution. Therefore, despite of considering model accuracy and complexity, we enforce the quantized networks to imitate the attribution of the full-precision counterparts. Instead of directly minimizing the Euclidean distance between attribution of quantized and full-precision models, we preserve their attribution rank consistency so that the attribution of quantized networks can adaptively adjust the distribution without capacity insufficiency. Figure \ref{comparison}(b) demonstrates the attribution computed by Grad-cam \cite{selvaraju2017grad} for mixed-precision networks with optimal and random quantization policy and their full-precision counterparts, where the mixed-precision networks with the optimal bitwidth assignment acquire more consistent attribution rank with the full-precision model. Experimental results show that our GMPQ obtains competitive accuracy-complexity trade-off on ImageNet and PASCAL VOC compared with the state-of-the-art mixed-precision quantization methods in only several GPU hours.

\section{Related Work}
\textbf{Fixed-precision quantization: }Network quantization has aroused extensive interests in computer vision and machine learning due to the significant reduction in computation and storage complexity, and existing methods are divided into one-bit and multi-bit quantization. Binary networks constrain the network weights and activations in one bit at extremely high compression ratio. For the former, Hubara \emph{et al.} \cite{hubara2016binarized} and Courbariaux \emph{et al.} \cite{courbariaux2016binarized} replaced the multiply-add operations with xnor-bitcount via weight and activation binarization, and applied the straight-through estimators (STE) to optimize network parameters. Rastegari \emph{et al.} \cite{rastegari2016xnor} leveraged the scaling factor for weight and activation hashing to minimize the quantization errors. Liu \emph{et al.} \cite{liu2018bi} added extra shortcut between consecutive convolutional layers to enhance the network capacity. Wang \emph{et al.} \cite{wang2019learning} mined the channel-wise interactions to eliminate inconsistent signs in feature maps. Qin \emph{et al.}  \cite{qin2020forward} minimized the parameter entropy in inference and utilized the soft quantization in backward propagation to enhance the information retention. Since the performance gap between full-precision and binary networks is huge, multi-bit networks are presented for better accuracy-efficiency trade-off. Zhu \cite{zhu2016trained} trained an adaptive quantizer for network ternarization according to weight distribution. Gong \emph{et al.} \cite{gong2019differentiable} applied the differentiable approximations for quantized networks to ensure the consistency between the optimization and the objective. Li \emph{et al.} \cite{li2019fully} proposed the four-bit networks for object detection with hardware-friendly implementations, and overcome the training instabilities by custom batch normalization and outlier removal. However, the fixed-precision quantization ignores the redundancy variance across different layers and leads to suboptimal accuracy-complexity trade-off in quantized networks.

\textbf{Mixed-precision quantization: }The mixed-precision networks assign different bitwidths to weights and activations in various layers, which considers the redundancy variance in different components to obatin the optimal accuracy-efficiency trade-off given hardware configurations. Existing mixed-precision quantization methods are mainly based on either non-differentiable or differentiable search. For the former, Wang \emph{et al.} \cite{wang2019haq} presented a reinforcement learning model to learn the optimal bitwidth for weights and activations of each layer, where the model accuracy and complexity were considered in reward function. Wang \emph{et al.} \cite{wang2020apq} jointly searched the pruning ratio, the bitwidth and the architecture of the lightweight model from a hypernet via the evolutionary algorithms. Since the non-differentiable methods require huge search cost to obtain the optimal bitwidth, the differentiable search approaches are also introduced in mixed-precision quantization. Cai \emph{et al.} \cite{cai2020rethinking} designed a hypernet where each convolutional layer consisted of parallel blocks in different bitwidths, which yielded the output by summing all blocks in various weights. Optimizing the block weight by back propagation and selecting the bitwidth with the largest value during inference achieved the optimal accuracy-complexity trade-off. Moreover, Yu \emph{et al.} \cite{yu2020search} further presented a barrier penalty to ensure that the searched models were within the complexity constraint. Yang \emph{et al.} \cite{yang2020automatic} decoupled the constrained optimization via Alternating Direction Method of Multipliers (ADMM), and Wang \emph{et al.} \cite{wang2020differentiable} utilized the variational information bottleneck to search for the proper bitwidth and pruning ratio. Habi \emph{et al.} \cite{habi2020hmq} and Van \emph{et al.} \cite{van2020bayesian} directly optimized the quantization intervals for bitwidth selection of mixed-precision networks. However, differentiable search for mixed-precision quantization still needs a large amount of time due to the optimization of the large hypernet. In order to solve this, Dong \emph{et al.} \cite{dong2019hawq, dong2019hawq-v2} designed bitwidth assignment rules according to Hessian information. Nevertheless, the hand-crafted rules require expert knowledge and cannot adapt to the input data.

\textbf{Attribution methods: }Attribution aims to produce human-understandable explanations for the predictions of neural networks. The contribution of each input component is calculated by examining the its influence on the network output, which is displayed as the attribution in 2D feature maps. Early works \cite{erhan2009visualizing,simonyan2013deep, zhou2016learning} analyzed the sensitivity and the significance of each pixel by leveraging its gradients with respect to the optimization objective. The recent studies on attribution extraction can be categorized into two types: gradient-based and relevance-based methods. For the first regard, Guided Backprop \cite{springenberg2014striving}, Grad-Cam \cite{selvaraju2017grad} and integrated gradient \cite{sundararajan2017axiomatic} combined the pixel gradients across different locations and channels for information fusion, so that more accurate attribution was obtained. For the latter regard, Zhang \emph{et al.} \cite{zhang2018top} constructed a hierarchical probabilistic model to mine the correlation between the input components and the prediction. In this paper, we observe that the attribution rank consistency of feature maps between vanilla and compressed networks benefits visual analysis for various data distribution, which is extended to generalizable mixed-precision quantization for significant search cost reduction.

\section{Approach}
In this section, we first introduce the mixed-precision quantization framework which suffers from significant search burden. Then we demonstrate the observation that the attribution rank consistency between full-precision and quantized models benefits visual analysis for various data distribution. Finally, we present the generalizable mixed-precision quantization via attribution rank preservation.

\begin{figure}[t]
	\centering
	\includegraphics[height=5.9cm, width=8cm]{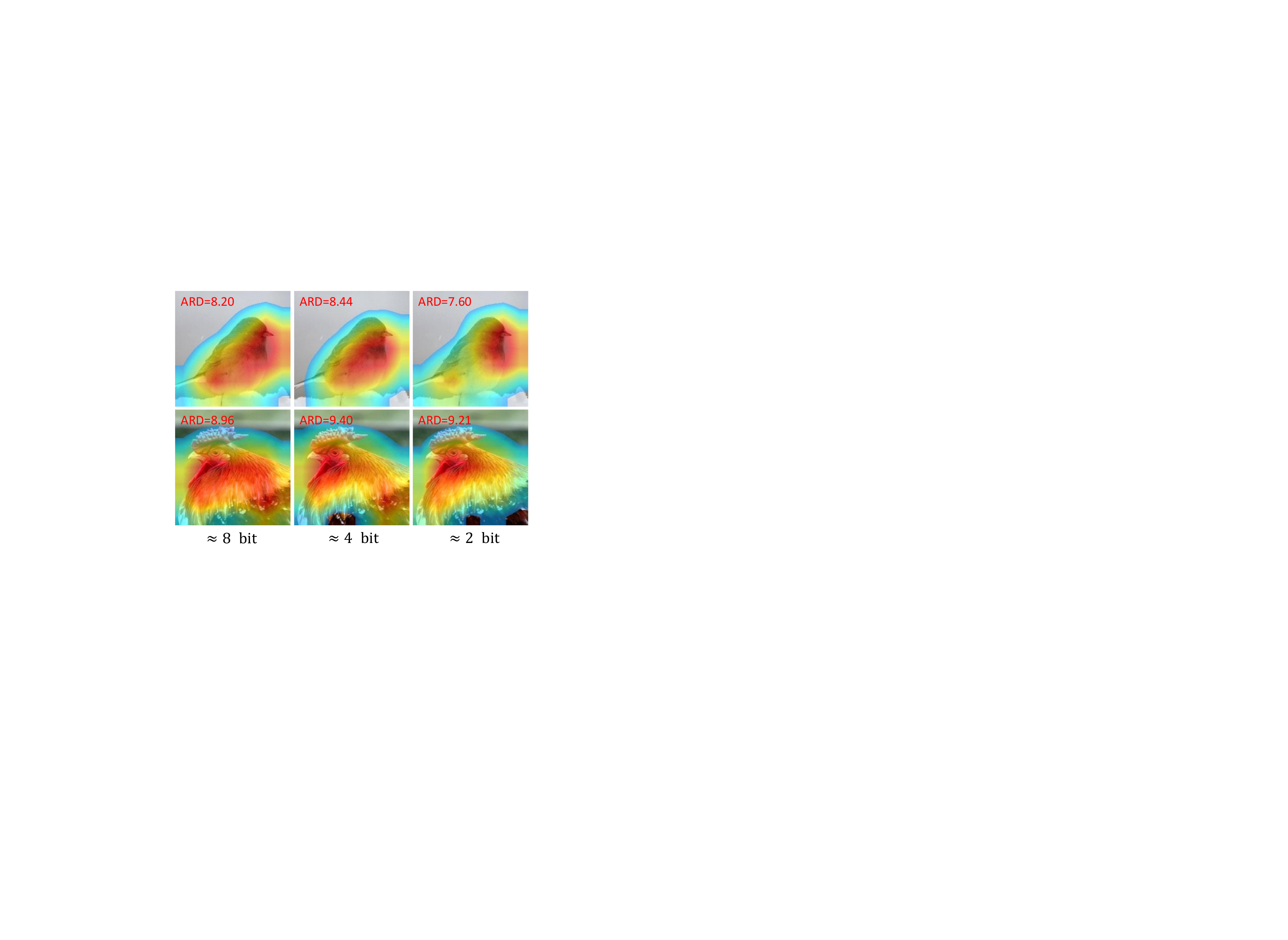}
	\caption{The attribution of the mixed-precision networks in different capacities with the optimal quantization policy. For the networks in low bitwidth, the attribution is more concentrated although the rank remains similar.  The concentrated attribution enables the model capacity to be sufficient by redundant attention removal, so that the promising performance is achieved.}
	\vspace{-0.5cm}    
	\label{complexity}
\end{figure}

\begin{figure*}[t]
	\centering
	\includegraphics[height=5.5cm, width=17cm]{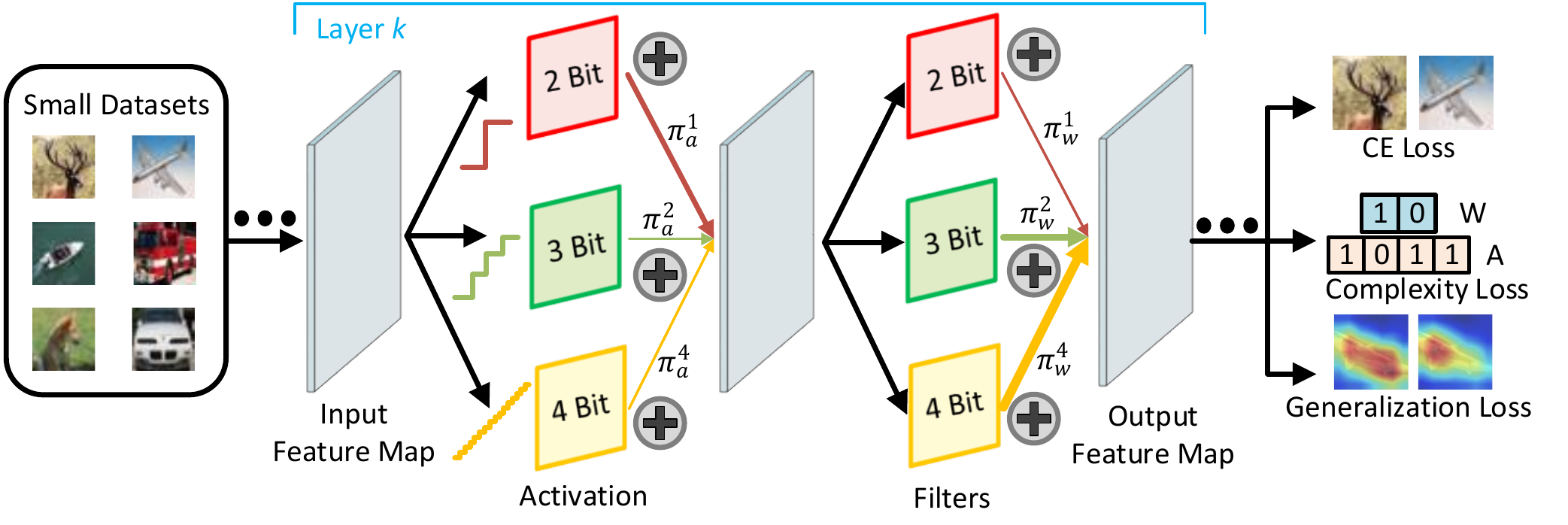}
	\caption{The pipeline of our GMPQ. The hypernet consists of multiple parallel branches including convolutional filters and activations in different bitwidths. The output from various branches is added with learnable importance weights to construct the output feature maps. Despite of the cross-entropy and complexity loss, we present additional generalization loss to optimize the network weights and branch importance weights, which enables the quantization policy searched on small datasets to be generalized on largescale datasets.}
	\vspace{-0.5cm}    
	\label{pipeline}
\end{figure*}

\subsection{Mixed-Precision Quantization}
The goal of mixed-precision quantization is to search the proper bitwidth of each layer in order to achieve the optimal accuracy-complexity trade-off given hardware configurations. Let $\bm{W}$ be the quantized network weight and $\mathcal{Q}$ be the quantization policy that assigns different bitwidths to weights and activations in various layers. $\Omega(\mathcal{Q})$ means the computational complexity of the compressed networks with the quantization policy $\mathcal{Q}$. The search objective function is written as the following bi-level optimization form:
\begin{align}
	\min\limits_{\mathcal{Q}}~~ &\mathcal{L}_{val}(\bm{W}^{*}(\mathcal{Q}), \mathcal{Q})\notag\\
	s.t. ~~&\bm{W}^{*}(\mathcal{Q})=\arg \min ~\mathcal{L}_{train}(\bm{W}, \mathcal{Q})\notag\\
	&\Omega(\mathcal{Q})\leqslant \Omega_0
\end{align}where $\mathcal{L}_{val}$ and $\mathcal{L}_{train}$ depict the task loss on the validation data and the training data. $\Omega_0$ stands for the resource constraint of the deployment platform. In order to obtain the optimal mixed-precision networks, the quantization policy $\mathcal{Q}$ and the network weights $\bm{W}(\mathcal{Q})$ are alternatively optimized until convergence or the maximal iteration number. Since the distribution of the training and validation data for policy search significantly affects the acquired quantization strategy, existing methods require the training and validation data for quantization policy search and those for model deployment to come from the same dataset. However, the compressed models are usually utilized on largescale datasets such as ImageNet, which causes heavy computational burden during quantization policy search. To address this, an ideal solution is to search for the quantization policy whose optimality is independent of the data distribution. The search objective should be modified in the following:
\begin{align}
	\min\limits_{\mathcal{Q}}~~&\mathbb{E}_{\bm{x}\sim\mathcal{D}_{val}^A}~\mathcal{L}(\bm{W}^{*}(\mathcal{Q}), \mathcal{Q}, \bm{x})\notag\\
	s.t. ~~&\bm{W}^{*}(\mathcal{Q})=\arg\min \mathbb{E}_{\bm{x}\sim \mathcal{D}_{train}^G} ~\mathcal{L}(\bm{W}, \mathcal{Q}, \bm{x})\notag\\
	&\Omega(\mathcal{Q})\leqslant \Omega_0
\end{align}where $\mathcal{L}(\bm{W}, \mathcal{Q}, \bm{x})$ represents the task loss for network weight $\bm{W}$, quantization policy $\mathcal{Q}$ and input $\bm{x}$. $\mathcal{D}_{val}^A$ depicts the dataset containing all validation images in deployment and $\mathcal{D}_{train}^G$ illustrates the dataset including given training images in bitwidth search, where the distribution gap between $\mathcal{D}_{val}^A$ and $\mathcal{D}_{train}^G$ may be sizable. Because $\mathcal{D}_{val}^A$ is intractable in realistic applications, it is desirable to find an alternative way to solve for the generalizable mixed-precision quantization policy.

\subsection{Attribution Rank Consistency}\label{Data-independent Attribution Consistency}
Since acquiring all validation images in deployment is impossible, we solve for the generalizable mixed-precision quantization policy via an alternative way. We observe that correctly locating the network attribution benefits visual analysis for various input data distribution. The feature attribution is formulated according to the loss gradient with respect to each feature map, where the importance of the $c_{th}$ feature map in the last convolutional layer for recognizing the objects from the $t_{th}$ class is written as follows:
\begin{align}\label{attention_wieght}
	\alpha_c[t]=\frac{1}{Z}\sum_{m,n}\frac{\partial f(\bm{x})[t]}{\partial A_c[m,n]}
\end{align}where $f(\bm{x})[t]$ means the output score for input $\bm{x}$ of the $t_{th}$ class, and $A_c[m,n]$ represents the activation element in the $m_{th}$ row and $n_{th}$ column of the $c_{th}$ feature map in the last convolutional layer. $Z$ is a scaling factor that normalizes the importance into the range $[0,1]$. With the feature map visualization techniques presented in Grad-cam \cite{selvaraju2017grad}, we obtain the feature attribution in the networks. We sum the feature maps from different channels with the attention weight calculated in (\ref{attention_wieght}), and remove the influence from opposite pixels via the ReLU operation. The feature attribution in the last convolutional layer with respect to the $t_{th}$ class is formulated in the following:
\begin{align}
\label{attribution}
M[t]=ReLU(\sum_{c}\alpha_c[t]\cdot \bm{A_c})
\end{align}The feature attribution only preserves the supportive features for the given class, and the negative features related to other classes are removed. 

The full-precision networks achieve high performance due to paying more attention to important parts in the image, while the quantized models deviate the attribution from that of the full-precision networks due to the limited capacity. Figure \ref{complexity} demonstrates the attribution of networks with the optimal quantization policy in different complexity, where attribution of networks in lower capacity is more concentrated due to the limited carried information. As the network capacity gap between the quantized networks and their full-precision counterparts is huge, directly enforcing the attribution consistency fails to remove the redundant attention in the compressed model, which causes capacity insufficiency with performance degradation. Therefore, we preserve the attribution rank consistency between the quantized networks and their full-precision counterparts for generalizable mixed-precision quantization policy search. The attribution rank illustrates the importance order of different pixels for model predictions. Constraining attribution rank consistency enables the quantized networks to focus on important regions, which adaptively adjusts the attribution distribution without capacity insufficiency.

\subsection{Generalizable Mixed-Precision Quantization via Attribution Rank Preservation}
Our GMPQ can be leveraged as a plug-and-play module for both non-differentiable and differentiable search methods. Since differentiable methods achieve the competitive accuracy-complexity trade-off compared with non-differentiable approaches, we employ the differentiable search framework \cite{cai2020rethinking, yu2020search, yang2020automatic} to select the optimal mixed-precision quantization policy. We design a hypernet with $N_a^k$ and $N_w^k$ parallel branches for convolution filters and feature maps in the $k_{th}$ layer. $N_a^k$ and $N_w^k$ represent the size of the search space for weight and activation bitwidths. The parallel branches are assigned with various bitwidths whose output is summed with the importance $\bm{\pi}_a^k$ and $\bm{\pi}_w^k$ for weight and activation respectively to form the intermediate feature maps. Figure \ref{pipeline} depicts the pipeline of our GMPQ. The feed-forward propagation for each layer in the $K$-layer hypernet is written as follows:
\begin{align}
	\bm{z}^k=\sum_{i=1}^{N_w^k}\pi_{w,i}^kf_i^k(\sum_{j=1}^{N_a^k}\pi_{a,j}^k\bm{a}_j^k)
\end{align}where $\bm{z}^k$ means the output intermediate feature maps of the $k_{th}$ layer. $\bm{a}_j^k$ represents the output of the $j_{th}$ activation quantization branch in the $k_{th}$ layer, and $f_i^k$ is the convolution operation in the $i_{th}$ filter branch of the $k_{th}$ layer. $\pi_{a,i}^k$ and $\pi_{w,i}^k$ stand for the importance weight for the $i_{th}$ quantized activation and filter branch in the $k_{th}$ layer. 

\begin{figure}[t]
	\centering
	\includegraphics[height=5.8cm, width=8cm]{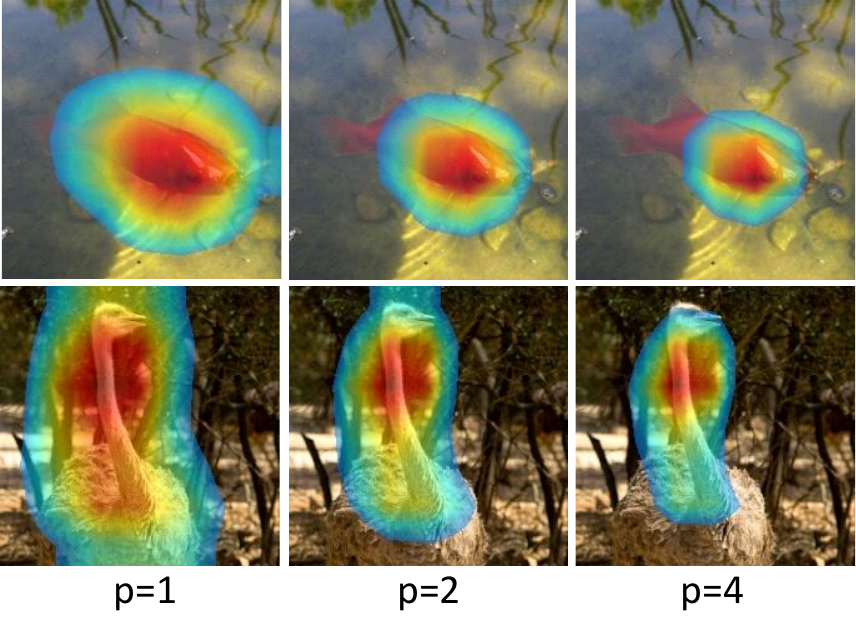}
	\caption{The $L_p$ norm of the attribution for the full-precision networks with different $p$. The attribution is more concentrated for larger $p$ while the rank keeps same.}
	\vspace{-0.5cm}    
	\label{pnorm}
\end{figure}

As we observe that the attribution rank consistency between quantized networks and their full-precision counterparts enables the compressed models to possess the discriminative power of the vanilla model regardless of the data distribution, we impose the attribution rank consistency constraint in optimal quantization policy search despite of the accuracy and efficiency objective. In order to obtain the optimal accuracy-complexity trade-off for generalizable mixed-precision quantization, the learning objective is formulated in the Lagragian form:
\begin{align}\label{risk}
	\mathcal{R}=\mathcal{R}_E(\bm{W},\mathcal{Q},\bm{x})+\zeta\mathcal{R}_C(\mathcal{Q})+\eta\mathcal{R}_G(\bm{W},\mathcal{Q},\bm{x})
\end{align}where $\mathcal{R}_E(\bm{W},\mathcal{Q},\bm{x})$,  $\mathcal{R}_C(\mathcal{Q})$ and $\mathcal{R}_G(\bm{W},\mathcal{Q},\bm{x})$ respectively mean the classification, complexity and the generalization risk for the networks with weight $\bm{W}$ and quantization policy $\mathcal{Q}$ for the input $\bm{x}$. $\zeta$ and $\eta$ are the hyperparameters to balance the importance of the complexity risk and generalization risk in the overall learning objective. In differentiable policy search, $\mathcal{R}_E(\bm{W},\mathcal{Q},\bm{x})$ is represented by the objective of vision tasks, and $\mathcal{R}_C(\mathcal{Q})$ is defined as the expected Bit-operations (BOPs) \cite{wang2020differentiable, bethge2020meliusnet, cai2020rethinking}:
\begin{align}
	\mathcal{R}_C(\mathcal{Q})=\sum_{k=1}^{K}(\sum_{i=1}^{N_w^k}\pi_{w,i}^kq_{w,i}^k)\cdot(\sum_{i=1}^{N_a^k}\pi_{a,i}^kq_{a,i}^k)\cdot B_{full}^k
\end{align}where $q_{w,i}^k$ and $q_{a,i}^k$ stand for the bitwidth of the $i_{th}$ branch of weights and activations in the $k_{th}$ layer, and $B_{full}^k$ means the BOPs of the $k_{th}$ layer in the full-precision network. $K$ represents the number of layers of the quantized model. As the attribution rank consistency between the full-precision networks and their quantized counterparts enhance the generalizability of the mixed-precision quantization policy, we define the generalization risk in the following form:
\begin{align*}
	\mathcal{R}_G(\bm{W},\mathcal{Q},\bm{x})=\sum_{i,j}||r(M_{q, ij}[y_x])-r(M_{f, ij}[y_x])||_2^2
\end{align*}where $M_{q, ij}[y_x]$ represents the pixel attribution in the $i_{th}$ row and $j_{th}$ column of the feature maps with respect to the class $y_x$ in the quantized networks, and $M_{f, ij}[y_x]$ demonstrates the corresponding variable in full-precision models. $y_x$ means the label of the input $\bm{x}$, and $||\cdot||_2$ is the element-wise $l_2$ norm. $r(\cdot)$ stands for the attribution rank, which equals to $k$ if the element is the $k_{th}$ largest in the attribution map. We only preserve the attribution rank consistency for top-k pixels with the highest attribution in the full-precision networks, as low attribution is usually caused by noise without clear information. Since minimizing the generalization risk is NP-hard, we present the capacity-aware attribution imitation to differentially optimize the objective. 

We enforce attribution of the mixed-precision networks to approach the $l_p$ norm of that in full-precision models, because the $l_p$ norm preserves the rank consistency while adaptively selects the attribution distribution according to the network capacity. The generalization risk is rewritten as follows for efficient optimization:
\begin{align*}
	\mathcal{R}_G(\bm{W},\mathcal{Q},\bm{x})=\sum_{i,j}||M_{q, ij}[y_x]-\frac{M_{f, ij}[y_x]^p}{\sum_{i,j}M_{f, ij}[y_x]^p}||_2^2
\end{align*}Large $p$ leads to concentrated attribution and vice versa, and we assign $p$ with larger value for hypernets in lower capacity with hyperparamters $Q_{w}^0$ and $Q_{a}^0$ for L-layer networks:
\begin{align}\label{capacity-aware}
	p=\frac{1}{L}\sum_{k=1}^{L}(Q_{w}^0/\sum_{i=1}^{N_w^k}\pi_{w,i}^kq_{w,i}^k)\cdot(Q_{a}^0/\sum_{i=1}^{N_a^k}\pi_{a,i}^kq_{a,i}^k)
\end{align}Since the classification, complexity and generalization risks are all differentiable, we optimize the hypernet weight and the branch importance weight iteratively in an end-to-end manner. When the hypernet converges or achieves the maximum training epoch, the bitwidth represented by the branch with the largest important weight is selected to form the final quantization policy. We finetune the quantized networks with the data in deployment to acquire the final model applied in realistic applications. GMPQ searches quantization policies on small datasets with generalization constraint, which leads to high performance on largescale datasets in deployment with significantly reduced search cost.

\section{Experiments}
In this section, we conducted extensive experiments on image classification and object detection. We first introduce the implementation details of our GMPQ. In the following ablation study, we then evaluated the influence of value assignment strategy for $p$ in the capacity-aware attribution imitation, investigated the effects of different terms in the risk function and discovered the impact of the dataset for quantization policy search. Finally, we compare our method with the state-of-the-art mixed-precision networks on image classification and object detection with respect to accuracy, model complexity and search cost.

\subsection{Datasets and Implementation Details}
We first introduce the datasets that we carried experiments on. For quantization policy search, we employed the small datasets including CIFAR-10 \cite{krizhevsky2009learning}, Cars \cite{krause20133d}, Flowers \cite{nilsback2008automated}, Aircraft \cite{maji2013fine}, Pets \cite{parkhi2012cats} and Food \cite{bossard2014food}. CIFAR-10 contains $60,000$ images divided into $10$ categories with equal number of samples, and Flowers have 8,189 images spread over 102 flower categories. Cars includes $16,185$ images with $196$ types at the level of maker, model and year, and Aircraft contains $10,200$ images with $100$ samples for each of the $102$ aircraft  model variants. Pet was created with $37$ dog and cat categories with $200$ images for each class, and Food contains $32,135$ high-resolution food photos of menu items from the $6$ restaurants. 

For mixed-precision network deployment, we evaluated the quantized networks on ImageNet for image classification and on PASCAL VOC for object detection. ImageNet \cite{deng2009imagenet} approximately contains $1.2$ billion and $50$k images for training and validation from $1,000$ categories. For training, $224\times224$ random region crops were applied from the resized image whose shorter side was $256$. During the inference stage, we utilized the $224\times224$ center crop. The PASCAL VOC dataset \cite{everingham2010pascal} collects images from $20$ categories, where we fintuned our mixed-precision networks on VOC 2007 and VOC 2012 trainval sets containing about $16$k images and tested our GMPQ on VOC 2007 test set consisting of $5$k samples. Following \cite{everingham2010pascal}, we used the mean average precision (mAP) as the evaluation metric. 

We trained our GMPQ with MobileNet-V2 \cite{sandler2018mobilenetv2}, ResNet18 and ResNet50 \cite{he2016deep} architectures for image classification, and applied VGG16 \cite{simonyan2014very} with SSD framework \cite{liu2016ssd} and ResNet18 with Faster R-CNN \cite{ren2015faster} for object detection. The bitwidth in the search space for network weights and activations is $2$-$8$ bit for MobileNet-V2 and $2$-$4$ bit for other architectures. Inspired by \cite{cai2020rethinking}, we utilized compositional convolution whose filters were weighted sum of each quantized filters in different bitwidths, so that complex parallel convolution was avoided. We updated the importance weight of different branches and the network parameters simultaneously. The hyperparameters $Q_w^0$ and $Q_a^0$ in capacity-aware attribution imitation were set to $4$ and $6$ respectively. Meanwhile, we only minimize the distance between attribution in quantized networks and $l_p$ norm of that in full-precision model for top-$1000$ pixels with the highest attribution in the real-valued model. For evaluation on ImageNet, we finetuned the mixed-precision networks with the Adam \cite{kingma2014adam} optimizer. The learning rate started from $0.001$ and decayed twice by multiplying $0.1$ at the $20_{th}$ and $30_{th}$ epoch out of the total $40$ epochs. For object detection, the backbone was pretrained on ImageNet and then finetuned on PASCAL VOC with the same hyperparameter settings on image classification. The batchsize was set to be $256$ in all experiments. By adjusting the hyperparameters $\zeta$ and $\eta$ in (\ref{risk}), we obtained the mixed-precision networks at different accuracy-complexity trade-offs. 

\begin{figure}[t]
	\centering
	\begin{center}
		\subfigure[Fixed strategy]{
			\begin{minipage}[b]{0.44\linewidth}\label{fig:5a}
				\includegraphics[height=3.5cm, width=4cm]{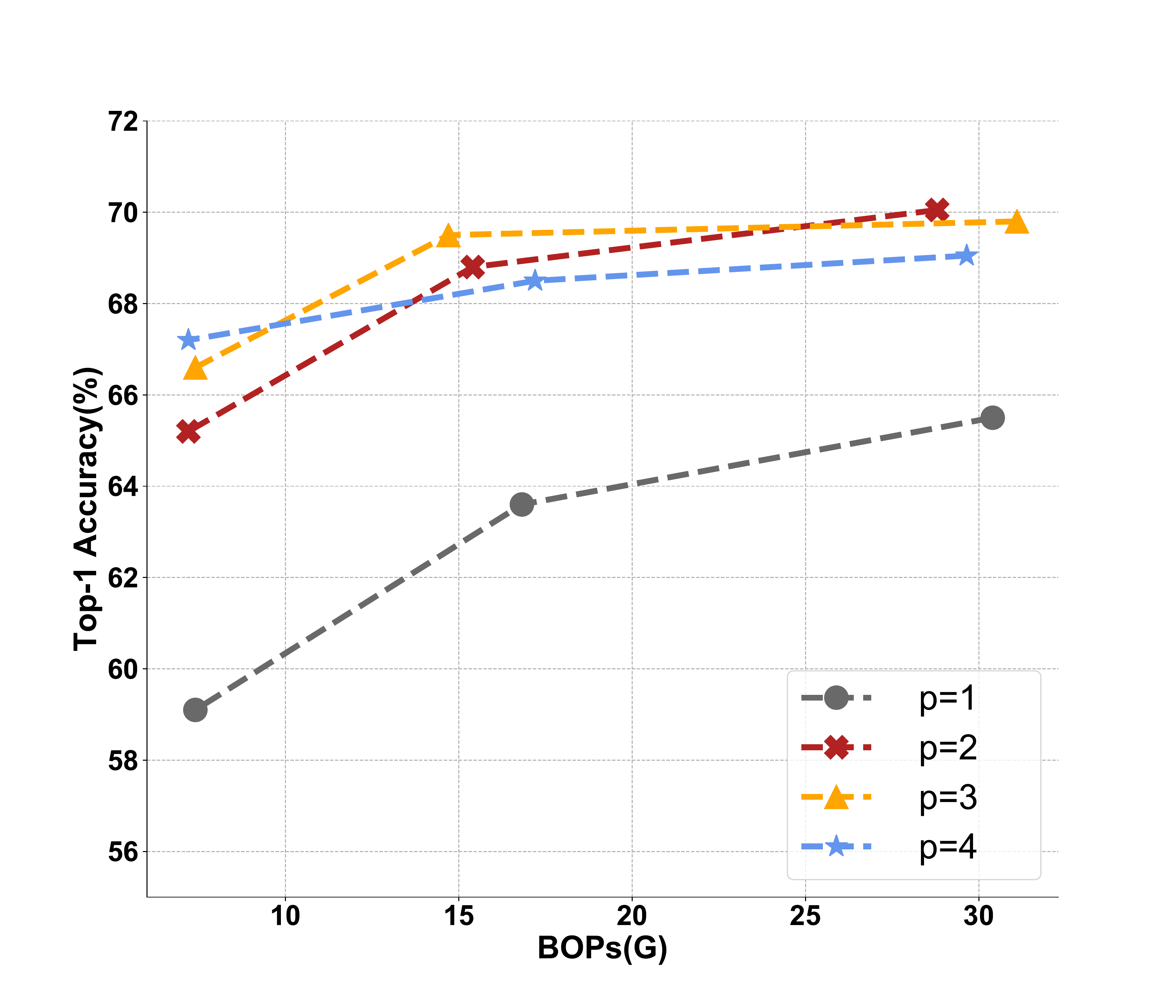}
			\end{minipage}
		}
		\hfill
		\subfigure[Capacity-aware strategy]{
			\begin{minipage}[b]{0.44\linewidth}\label{fig:5b}
				\includegraphics[height=3.5cm, width=4cm]{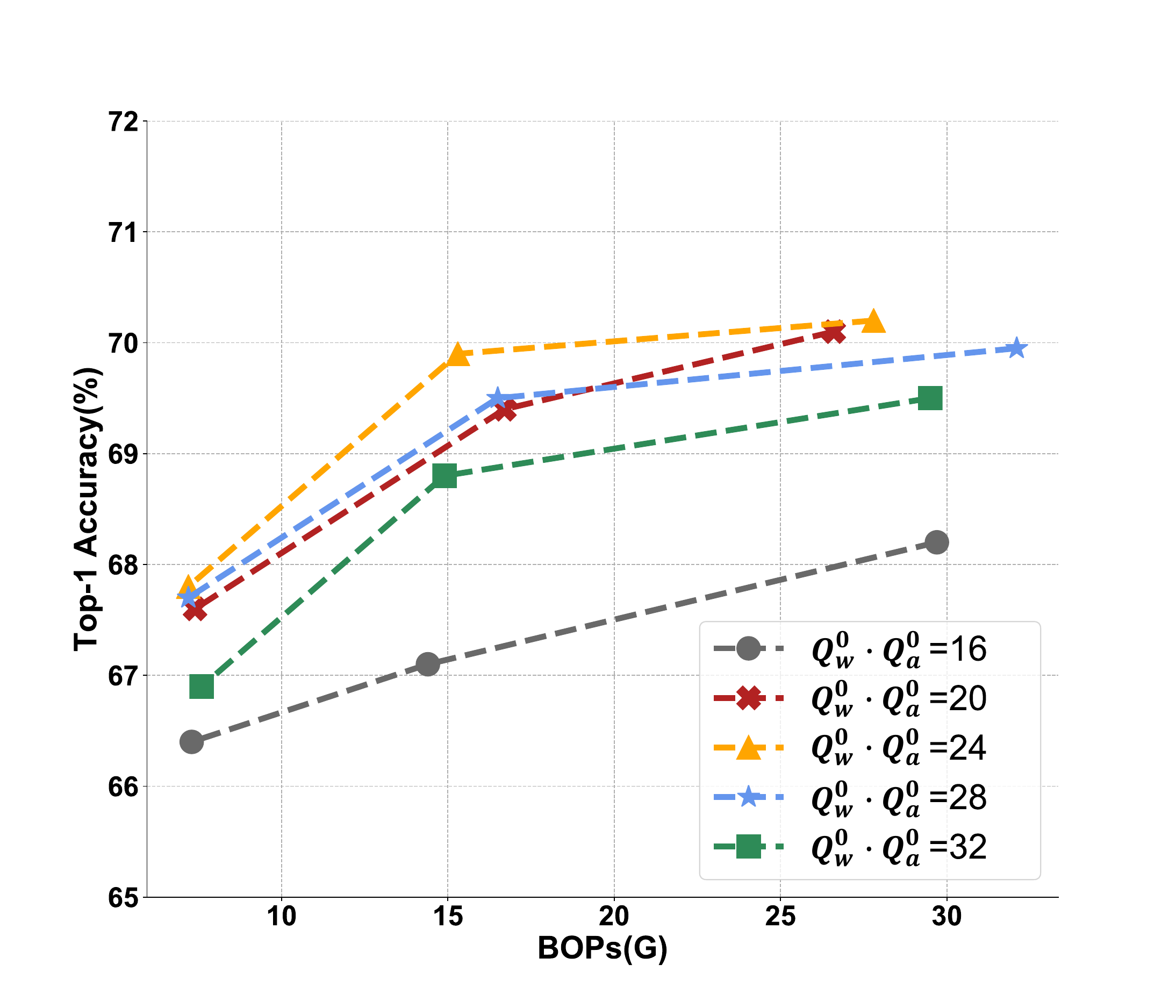}
			\end{minipage}
		}
		\vspace{-0.1cm}
		\caption{The accuracy-complexity trade-off of (a) fixed and (b) capacity-aware value assignment strategies for $p$ in (\ref{capacity-aware}), where hyperparameters are also varied.}
		\vspace{-0.3cm}
	\end{center}
	\label{ablation}
	\vspace{-0.5cm}
\end{figure}

\subsection{Ablation Study}
In order to investigate the effectiveness of attribution rank preservation, we assign the value of $p$ in the capacity-aware attribution imitation with different strategies. By varying the hyperparameters $\zeta$ and $\eta$ in the overall risk (\ref{risk}), we evaluated the influence of classification, complexity and generalization risks with respect to the model accuracy and efficiency. We conducted the ablation study on ImageNet with the ResNet18 architecture, and searched the mixed-precision quantization policy on CIFAR-10 for the above investigation. Moreover, we searched the generalizable mixed-precision quantization policy on different small datasets to discover the effects on the accuracy-complexity trade-off and search cost.

\begin{figure}[t]
	\centering
	\begin{center}
		\subfigure[Varying $\zeta$ and $\eta$]{
			\begin{minipage}[b]{0.44\linewidth}\label{fig:6a}
				\includegraphics[height=3.5cm, width=4cm]{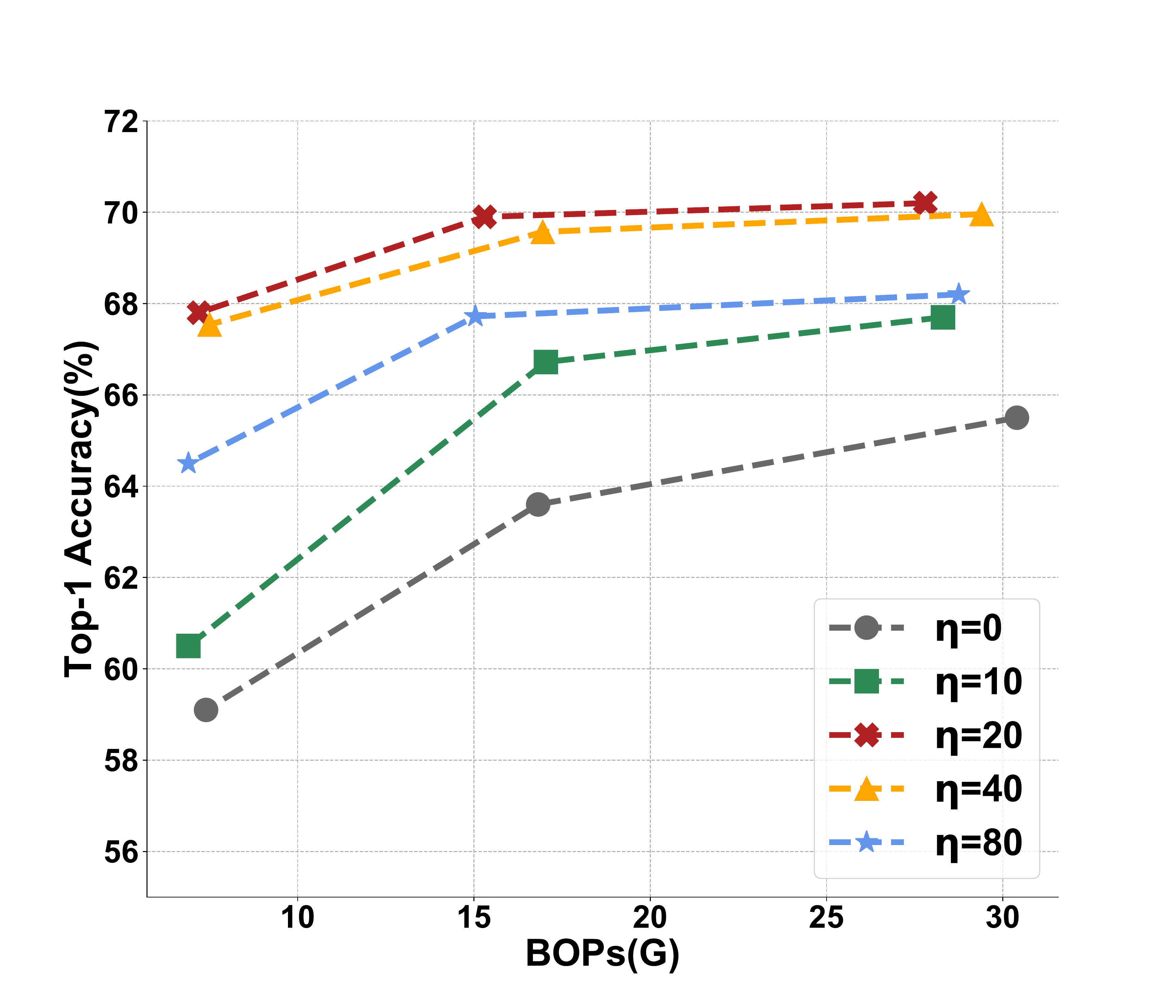}
			\end{minipage}
		}
		\hfill
		\subfigure[Varying datasets]{
			\begin{minipage}[b]{0.44\linewidth}\label{fig:6b}
				\includegraphics[height=3.5cm, width=4cm]{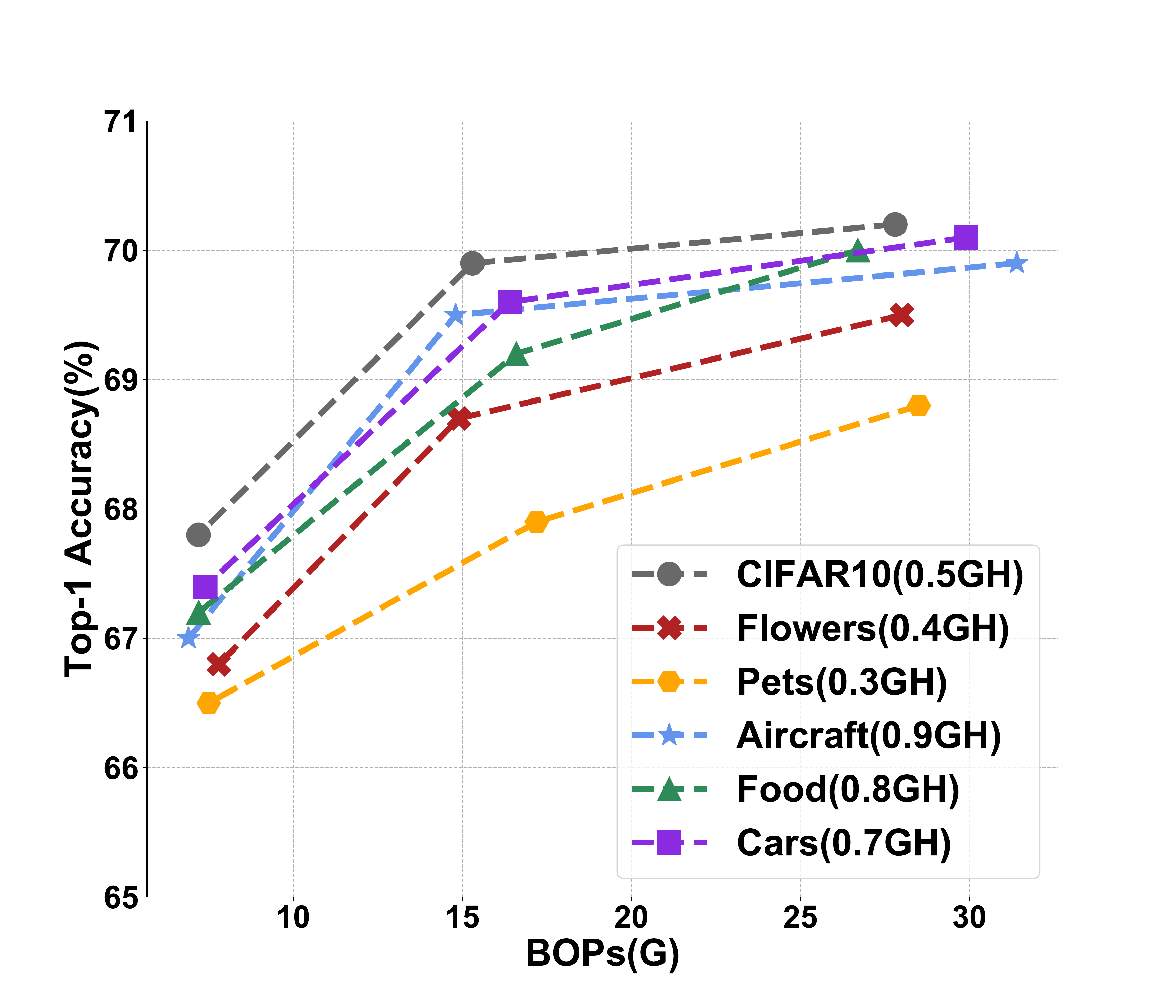}
			\end{minipage}
		}
		\vspace{-0.1cm}
		\caption{(a) The accuracy-complexity trade-off for different $\eta$, where $\zeta$ is varied to select various network capacity. (b) The top-1 accuracy on ImageNet, the BOPs and the average search cost of the mixed-precision quantization policy searched on different small datasets, where GH means GPU hours for the search cost.}
		\vspace{-0.3cm}
	\end{center}
	\label{dataset}
	\vspace{-0.5cm}
\end{figure}

\textbf{Effectiveness of different value assignment strategies for $p$: }To investigate the influence of value assignment strategies to $p$ on the accuracy-complexity trade-off, we searched the mixed-precision quantization policy with fixed and capacity-aware $p$ value. For fixed $p$, we set the value as $1$, $2$, $3$ and $4$ that constrains the attribution of quantized networks with various concentration. The capacity-aware strategy assigns $p$ with the strategy shown in (\ref{capacity-aware}), where the product of $Q_w^0$ and $Q_a^0$ was varied in the ablation study. Figure \ref{fig:5a} and \ref{fig:5b} demonstrate the accuracy-complexity trade-off for fixed and capacity-aware value assignment strategies for $p$ respectively with different hyperparameters. The optimal accuracy-complexity curve in capacity-aware strategy outperforms that in fixed strategy, which indicates the importance of attribution variation with respect to network capacity. For fixed strategy, medium $p$ outperforms other values. Small $p$ causes attention redundancy for quantized networks with limited capacity and large $p$ leads to information loss that fails to utilize the network capacity. For capacity-aware strategy, setting the product of $Q_w^0$ and $Q_a^0$ to $24$ results in the optimal accuracy-complexity trade-off. For hypernetworks whose product of weight and activation bitwidths is $24$, the network capacity is comparable with their full-precision counterparts since they mimic the attribution of real-valued models without extra concentration.

\begin{table}[t]
	\renewcommand\arraystretch{1.1}
	\footnotesize
	\caption{The top-1/top-5 accuracy (\%) on ImageNet, model storage cost (M),  model computational cost (G) and the search cost (GPU hours) for networks in different capacity and mixed-precision quantization policy. Param. means the model storage cost, and Comp. means the compression ratio of BOPs. }
	\label{ImageNet}
	\centering
	\vspace{0.1cm}
	\renewcommand\arraystretch{1.1}
	\begin{tabular}{p{1.6cm}<{\centering}|p{0.6cm}<{\centering}p{0.6cm}<{\centering}p{0.6cm}<{\centering}|p{0.7cm}<{\centering}p{0.7cm}<{\centering}|p{0.5cm}<{\centering}}
		\hline
		Methods & Param. & BOPs & Comp. & Top-1 & Top-5 & Cost. \\
		\hline
		\hline
		\multicolumn{7}{c}{ResNet18}\\
		\hline
		Full-precision & $46.8$ &$1853.4$ &$-$& $69.7$ & $89.2$  & $-$\\
		\hline 
		ALQ & $1.8$&$58.5$&$31.7$ & $67.7$ & $-$  & $34.7$\\
		HAWQ & $5.8$&$34.0$&$54.5$ & $68.5$ & $-$ & $15.6$\\
		GMPQ & $5.4$&$27.8$&$66.7$ & $70.2$ & $90.1$  & $0.5$\\
		\hline
		APoT & $4.6$& $16.3$& $113.8$& $69.8$ & $-$  & $-$\\
		GMPQ & $4.1$& $15.3$& $121.0$ & $69.9$ & $89.7$  & $0.6$\\
		\hline
		ALQ & $3.4$ &$7.2$&$256.0$ & $66.4$ & $-$  & $38.5$\\
		EdMIPS& $4.7$&$7.2$&$258.0$ & $65.9$ & $86.5$  & $9.5$\\
		EdMIPS-C & $4.5$&$7.4$&$251.9$ & $59.1$ & $81.0$  & $0.6$\\
		GMPQ & $3.7$&$7.2$&$255.8$ & $67.8$ & $88.0$  & $0.9$\\
		\hline
		\hline
		\multicolumn{7}{c}{ResNet50}\\
		\hline
		Full-precision & $97.5$&$3952.6$ &$-$& $76.4$ & $93.1$  & $-$\\
		\hline 
		HAWQ & $13.1$&$61.3$&$64.5$ & $75.3$ & $92.4$  & $36.6$\\
		HAQ & $12.2$&$50.3$&$78.6$ & $75.5$ & $92.4$  & $67.7$\\
		BP-NAS & $13.4$&$55.2$&$71.7$ & $76.7$ & $93.6$  & $30.2$\\
		GMPQ & $12.4$&$53.0$&$74.6$ & $76.7$ & $93.3$  & $2.2$\\
		
		\hline
		HMQ& $15.6$&$37.7$&$104.8$ & $75.5$ & $-$  & $49.4$\\
		BP-NAS& $11.3$&$33.2$&$119.0$ & $75.7$ & $92.8$  & $35.6$\\
		GMPQ & $9.6$&$30.7$&$128.6$ & $75.8$ & $92.9$  & $2.7$\\
		\hline
		EdMIPS& $13.9$&$15.6$&$254.2$ & $72.1$ & $90.6$  & $26.5$\\
		EdMIPS-C & $13.7$&$16.0$&$247.2$ & $65.6$ & $87.3$  & $2.9$\\
		GMPQ & $8.8$&$15.7$&$252.2$ & $73.6$ & $91.2$  & $3.4$\\
		\hline 
		\hline 
		\multicolumn{7}{c}{MobileNet-V2}\\
		\hline
		Full-precision & $13.4$&$337.9$&$-$ & $71.9$ & $90.3$  & $-$\\
		\hline
		RQ & $2.7$&$11.9$&$28.4$ & $68.0$ & $-$  & $-$\\
		GMPQ & $1.4$&$10.4$&$32.6$ & $71.5$ & $90.2$  & $1.7$\\
		\hline 
		HAQ   & $1.4$&$8.3$&$41.0$ & $69.5$ & $88.8$  & $51.1$\\
		HAQ-C & $1.6$&$8.1$&$41.6$ & $62.7$ & $82.4$  & $4.5$\\
		DJPQ & $1.9$&$7.9$&$43.0$ & $69.3$ & $-$  & $12.2$\\
		GMPQ  & $1.2$&$7.4$&$45.8$ & $70.4$ & $90.7$  & $2.6$\\
		
		\hline
		HMQ & $1.7$&$5.2$&$64.4$ & $70.9$ & $-$  & $33.5$\\
		DQ & $1.7$&$4.9$&$68.7$ & $69.7$ & $-$  & $21.6$\\
		GMPQ & $1.0$&$4.8$&$69.7$ & $70.1$ & $89.9$  & $2.8$\\
		\hline
	\end{tabular}
	\vspace{-0.7cm}
\end{table}

\textbf{Influence of hyperparameters in overall risk (\ref{risk}): }In order to verify the effectiveness of the generalization risk, we report the performance with different $\eta$. Meanwhile, we also varied the hyperparameter $\zeta$ to obtain different accuracy-complexity trade-offs. Figure \ref{fig:6a} illustrates the results, where medium $\eta$ achieves the best trade-off curve. Large $\eta$ fails to leverage the supervision from annotated labels, and small $\eta$ ignores the attribution rank consistency which enhances the generalization ability of the mixed-precision quantization policy. With the increase of $\zeta$, the resulted policy prefers lightweight architectures and vice versa. For different $\eta$, the same assignment of $\zeta$ selects similar BOPs in the accuracy-complexity trade-off. 

\textbf{Effects of datasets for quantization policy search:} We searched the mixed-precision quantization policy on different small datasets including CIFAR-10, Cars, Flowers, Aircraft, Pets and Food to discover the effects on model accuracy and efficiency. Figure \ref{fig:6b} demonstrates the top-1 accuracy and the BOPs for the optimal mixed-precision networks obtained on different small datasets. We also show the average search cost across all computation cost constraint in the legend, where GH means GPU hours that measures the search cost. The mixed-precision networks searched on CIFAR-10 achieves the best accuracy-efficiency trade-off, because the size of CIFAR-10 is the largest with the most sufficient visual information. Moreover, the gap of object category between CIFAR-10 and ImageNet is the smallest compared with other datasets. Searching quantization policy on Aircraft requires the highest search cost due to the large image size $512\times512$. 

\begin{table}[t]
	\footnotesize
	\caption{The mAP (\%) on PASCAL VOC, model storage cost (M),  model computational cost (G) and the search cost (GPU hours) for backbone networks in different capacity and mixed-precision quantization policy. Param. means the model storage cost, and Comp. means the compression ratio of BOPs. }
	\label{VOC}
	\centering
	\vspace{0.1cm}
	\renewcommand\arraystretch{1.1}
	\begin{tabular}{p{1.6cm}<{\centering}|p{0.8cm}<{\centering}p{0.8cm}<{\centering}p{0.8cm}<{\centering}|p{0.8cm}<{\centering}p{0.8cm}<{\centering}}
		\hline
		Methods & Param. & BOPs & Comp. & mAP  & Cost\\
		\hline
		\hline
		\multicolumn{6}{c}{SSD \& VGG16}\\
		\hline
		Full-precision & $105.5$ & $27787.7$ & $-$ & $72.4$ &$-$ \\
		\hline
		HAQ & $42.7$&$847.2$ & $32.8$ & $70.9$ &$62.5$ \\
		HAQ-C & $42.9$&$819.7$ & $33.9$ & $67.6$ &$5.1$ \\
		EdMIPS& $33.5$&$958.2$ & $29.0$ & $69.4$ &$25.9$ \\
		EdMIPS-C& $37.2$&$868.4$ & $32.0$ & $65.2$ &$1.5$ \\
		GMPQ & $36.6$&$796.2$ & $34.9$ & $70.5$ &$1.6$ \\
		\hline 
		HAQ      & $35.5$ &  $430.15$ & $64.6$ & $69.1$ &$67.9$ \\
		HAQ-C 	 & $32.3$ &  $445.3$ & $62.4$ & $66.4$ &$6.8$ \\
		EdMIPS   & $29.4$ & $454.0$ & $61.2$ & $68.7$ &$30.2$ \\
		EdMIPS-C & $31.3$ & $423.6$ & $65.6$ & $64.3$ &$1.6$ \\
		GMPQ     & $24.7$&$413.5$ & $67.2$ & $69.2$ &$1.8$ \\
		
		\hline
		\hline
		\multicolumn{6}{c}{Faster R-CNN \& ResNet18}\\
		\hline
		Full-precision &$47.4$& $22534.8$& $-$ & $74.5$ &$-$  \\
		\hline 
		HAQ & $8.3$&$342.5$ & $65.8$ & $73.5$ &$38.9$  \\
		HAQ-C & $8.5$&$337.9$ & $66.7$ & $70.7$ &$4.1$  \\
		EdMIPS& $9.3$&$361.7$ & $62.3$ & $72.3$ &$16.6$  \\
		EdMIPS-C & $8.7$&$348.8$ & $64.6$ & $69.8$ &$0.4$  \\
		GMPQ & $6.4$&$337.9$ & $66.7$ & $73.9$ & $0.5$  \\
		\hline
		HAQ & $8.0$&$303.7$ & $74.2$ & $73.2$ &$35.2$  \\
		HAQ-C & $7.6$&$310.4$ & $72.6$ & $70.4$ &$5.2$  \\
		EdMIPS& $18.7$&$348.8$ & $71.1$ & $71.8$ &$18.1$  \\
		EdMIPS-C & $7.4$&$299.3$ & $75.3$ & $69.2$ &$0.4$  \\
		GMPQ & $6.2$&$286.3$ & $78.7$ & $73.4$ & $0.5$ \\
		\hline 
	\end{tabular}
	\vspace{-0.5cm}
\end{table}

\subsection{Comparison with State-of-the-art Methods}
In this section, we compare our GMPQ with the state-of-the-art fixed-precision models containing APoT \cite{li2019additive} and RQ \cite{louizos2018relaxed} and mixed-precision networks including ALQ \cite{qu2020adaptive}, HAWQ \cite{dong2019hawq}, EdMIPS \cite{cai2020rethinking}, HAQ \cite{wang2019haq}, BP-NAS \cite{yu2020search}, HMQ \cite{habi2020hmq} and DQ \cite{uhlich2019differentiable} on ImageNet for image classification and on PASCAL VOC for object detection. We also provide the performance of full-precision models for reference. The accuracy-complexity trade-offs of baselines are copied from their original papers or obtained by our implementation with the officially released code, and the search cost was evaluated by re-running the compared methods. We searched the optimal quantization policy on CIFAR-10 for the deployment on ImageNet and PASCAL VOC.

\textbf{Results on ImageNet: }Table \ref{ImageNet} illustrates the comparison of storage and computational cost, the compression ratio of BOPs, the top-1 and top-5 accuracy and the search cost across different architectures and mixed-precision quantization methods.  HAQ-C and EdMIPS-C demonstrate that we leveraged HAQ and EdMIPS that searched the quantization policy on CIFAR-10 and evaluated the obtained quantization policy on ImageNet. By comparing the accuracy-complexity trade-off with the baseline methods for different architectures, we conclude that our GMPQ achieves the competitive accuracy-complexity trade-off under various resource constraint with significantly reduced search cost. Meanwhile, we also searched the quantization policy on CIFAR-10 directly using HAQ and EdMIPS. Although the search cost is reduced sizably, the accuracy-complexity trade-off is far from the optimal across various resource constraint, which indicates the lack of generalization ability for the quantization policy obtained by the conventional methods. Our GMPQ preserves the attribution rank consistency during the quantization policy search with acceptable computational overhead, and enables the mixed-precision quantization searched on small datasets to generalize to largescale datasets. For the mixed-precision quantization method EdMIPS, the search cost reduction is more obvious for ResNet50 since the heavy architecture requires more training epochs to converge when trained on largescale datasets.

\textbf{Results on PASCAL VOC: }We employed the SSD detection framework with VGG16 architecture and Faster R-CNN detector with ResNet18 backbone to evaluate our GMPQ on object detection. Table \ref{VOC} shows the results of various mixed-precision networks. Compared with the accuracy-complexity trade-off searched on PASCAL VOC by the state-of-the-art methods, our GMPQ acquired the competitive results with significantly reduced search cost on both detection frameworks and backbones. Moreover, directly compressing the networks with the quantization policy searched by HAQ and EdMIPS on CIFAR-10 degrades the performance significantly. Since the mixed-precision networks are required to be pretrained on ImageNet, the search cost decrease on PASCAL VOC is more sizable than that on ImageNet. Because the two-stage detector Faster R-CNN has stronger discriminative power for accurate attribution generation, the accuracy-complexity trade-off is more optimal compared with the one-stage detector.

\section{Conclusion}
In this paper, we have proposed a generalizable mixed-quantization method called GMPQ for efficient inference. The presented GMPQ searches the quantization policy on small datasets with attribution rank preservation, so that the acquired quantization strategy can be generalized to achieve the optimal accuracy-complexity trade-off on largescale datasets with significant search cost reduction. Extensive experiments depict the superiority of GMPQ compared with the state-of-the-art methods.

\section*{Acknowledgements}
This work was supported in part by the National Key Research and Development Program of China under Grant 2017YFA0700802, in part by the National Natural Science Foundation of China under Grant 61822603, Grant U1813218, and Grant U1713214, in part by a grant from the Beijing Academy of Artificial Intelligence (BAAI), and in part by a grant from the Institute for Guo Qiang, Tsinghua University.

{\small
	\bibliographystyle{ieee_fullname}
	\bibliography{egbib}
}

\end{document}